\documentclass{article} % For LaTeX2e
\usepackage{iclr2021_conference,times}

\usepackage{amsmath,amsfonts,bm}

\def\eqref#1{equation~\ref{#1}}
\def\1{\bm{1}}

\DeclareMathAlphabet{\mathsfit}{\encodingdefault}{\sfdefault}{m}{sl}
\SetMathAlphabet{\mathsfit}{bold}{\encodingdefault}{\sfdefault}{bx}{n}

\usepackage{hyperref}
\usepackage{url}
\usepackage{graphicx}
\usepackage{subcaption}
\usepackage{floatrow}
\newfloatcommand{capbtabbox}{table}[][\FBwidth]

\title{Ensembles of GANs for synthetic training data generation}

\author{Gabriel Eilertsen$^{1,2}$,\; Apostolia Tsirikoglou$^1$,\; Claes Lundstr\"om$^{1,2,3}$,\; Jonas Unger$^{1,2}$
\vspace{0.2cm}
\\
$^1$~Department of Science and Technology, Linköping University, Sweden \\
$^2$~Center for Medical Image Science and Visualization, Linköping University, Sweden \\
$^3$~Sectra AB, Sweden  \\
}

\newcommand{\heading}[1]{\section{#1}\vspace{-0.2cm}}
\vspace{-0.2cm}

\iclrfinalcopy % Uncomment for camera-ready version, but NOT for submission.
\begin{document}

\maketitle
\vspace{-0.4cm}

\begin{abstract}
\vspace{-0.15cm}
Insufficient training data is a major bottleneck for most deep learning practices, not least in medical imaging where data is difficult to collect and publicly available datasets are scarce due to ethics and privacy. This work investigates the use of synthetic images, created by generative adversarial networks (GANs), as the only source of training data. We demonstrate that for this application, it is of great importance to make use of multiple GANs to improve the diversity of the generated data, i.e. to sufficiently cover the data distribution. While a single GAN can generate seemingly diverse image content, training on this data in most cases lead to severe over-fitting. We test the impact of ensembled GANs on synthetic 2D data as well as common image datasets (SVHN and CIFAR-10), and using both DCGANs and progressively growing GANs. As a specific use case, we focus on synthesizing digital pathology patches to provide anonymized training data.
\end{abstract}

\heading{Introduction}
A deficiency of training data is often limiting the performance of deep learning models, especially in areas such as medical imaging, where acquisition and annotation is highly time-consuming and reliant on busy experts. One potential solution is to use generative techniques to synthesize training data. 
%If high-quality data collections could be generated, it would greatly lower the threshold for creating accurate deep learning applications. 
%Such anonymization at scale would be a quantum leap for putting medical data to use in AI development.
%Generative adversarial networks (GANs) have shown great success for learning-based image synthesis~\citep{Goodfellow2014}. With recent advances, GANs can produce images difficult to distinguish from real, given that the domain under consideration is sufficiently narrow (e.g. faces, letters, dogs, etc.)~\citep{Karras2018,Karras2019}, and even for separate classes in class-conditioned generation of more complicated datasets~\citep{Brock2018,Miyato2018,Zhang2019}.
The high quality of images from generative adversarial networks (GANs)~\citep{Goodfellow2014,Karras2019,Brock2018,Miyato2018,Zhang2019} have proven effective for image augmentation in deep learning for medical imaging applications~\citep{Frid2018,Madani2018,Bowles2018}. 
Moreover, 
%as access to healthcare data is strongly limited due to patient integrity concerns, 
the possibility of creating purely synthetic collections representative of non-shareable patient data would be an effective anonymization approach~\citep{Guibas2017,Shin2018,Triastcyn2019,Yoon2020}.

%Despite the recent success of GANs, a remaining fundamental challenge is to ensure and control that the underlying data distribution is sufficiently covered. This is especially problematic when attempting to only use synthetic data in training. The typical pitfall is mode collapse, that the objective of the optimization is satisfied if the main modes of the distribution are reproduced, even though other less common features are ignored. Although data diversity can explicitly be included in the loss~\citep{Karras2018}, full coverage is not ensured. For example, recent studies show how GANs can completely skip certain image content while still generating high-quality images~\citep{Bau2019}. From the perspective of using GANs to produce training data, this is a serious problem, as the under-represented samples can contribute substantially to the performance of the end application. 

We focus on the impact of using GANs in ensembles for the purpose of synthesizing training data for downstream deep learning applications. In particular, we are interested in the scenario where releasing the real data is not an option, e.g. if this is of private/sensitive nature or in other ways protected. This problem formulation differs significantly from considering training on real and synthetic data in combination, i.e. using GANs as a tool for data augmentation, where the real data can carry the bulk of information while the synthetic data provide some variations to improve generalization. Requiring the synthetic data to carry all information makes diversity a critical aspect, and missing modes of the underlying data distribution can have a large influence. %heavily influence the quality of the downstream application.
%Although data diversity can explicitly be included in the loss~\citep{Karras2018}, full coverage is not ensured. For example, recent studies show how GANs can completely skip certain image content while still generating high-quality images~\citep{Bau2019}.
Previous work has focused on techniques for increasing the mode coverage of GANs~\citep{Lin2018,Hoang2018,Liu2020}. A powerful solution is to train ensembles of multiple GANs~\citep{wang2016ensembles,Tolstikhin2017,Grover2018}, and boosting strategies have been considered for forcing the GANs to focus on different parts of the data distribution. While boosting makes intuitive sense, we show that for the purpose of generating training data the most important aspect is to combine multiple GANs; these could even be trained completely independently, meaning that the stochasticity of the optimization is powerful enough to allow for good mode coverage. %To illustrate this argument, Figure~\ref{fig:2dgans} shows the limited mode coverage of individually trained GANs on a 2D dataset drawn from a mixture of Gaussians. When combining multiple models, the mode coverage improves quickly, and for this dataset 2-3 models is enough to ensure full mode coverage.

%The contributions of this work are three-fold:
Our contributions include:
\textbf{1)} we perform an evaluation of the behavior of deep classifiers when trained on purely GAN generated training data,
\textbf{2)} we highlight the importance of using ensembles of GANs for synthesizing datasets diverse enough for training deep models, and show that independently trained GANs could be advantageous compared to boosting strategies, with as good or better performance and less prone to over-fitting, and
\textbf{3)} we test ensemble performance on 2D data, on SVHN and CIFAR-10, as well as in a more realistic scenario of anonymization in digital pathology.

\heading{Ensemble GANs}
A GAN generator $G(z)$ produces synthetic image samples by drawing from the latent distribution $z \sim \mathcal{Z}$. 
%$G$ is trained with the objective to decrease the accuracy of a discriminator $D(x)$ that is trained to distinguish between real images $x$ and generated images
We define an ensemble of $T$ GANs as the mixture $\hat{G}_T = \sum_{t=1}^Tp_tG_t$, 
%
%\begin{equation}
%\hat{G}_T = \sum_{t=1}^Tp_tG_t,
%\label{eqn:ensemble}
%\end{equation}
%
i.e. a weighted combination of individually trained GANs $G_t$, where $\sum_t p_t = 1$. In practice, generating images from $\hat{G}_T$ entails drawing individual samples from $G_t$ with probability $p_t$. This means that a generated dataset of $N_g$ samples will contain $p_t N_g$ samples from $G_t$.
In practice, we consider classification as the downstream application for the synthetic data. This means that we have access to class labels $k \in [1,K]$. To make use of this information, we use bootstrap aggregation (bagging), by training GANs separately for each class $k$ in each ensemble iteration. Given that we strive for maximal diversity and mode coverage, it is a sensible choice to reflect the different parts of the data distribution covered by the individual classes of the dataset.
%We have opted for this approach instead of alternatives such as class-conditioned GANs that could control generation of samples from different classes in one single model.
This means that in each ensemble iteration we train $K$ models $G_{t,k}$, and the total mixture is $\hat{G}_T = \frac{1}{K}\sum_{k=1}^{K}\sum_{t=1}^Tp_tG_{t,k}$.
%
%\begin{equation}
%\hat{G}_T = \frac{1}{K}\sum_{k=1}^{K}\sum_{t=1}^Tp_tG_{t,k}.
%\label{eqn:ensemble}
%\end{equation}
%
Training a separate GAN for each class means, in practice, that we will have an ensemble/mixture of $KT$ GANs. However, for better clarity we will only refer to the iterations $T$.

%\paragraph{Independent trainings vs. boosting}
We are mainly interested in independent ensembles, where $G_t$ are trained in isolation from each other. This means that the benefit of combining multiple GANs comes entirely from the stochastic nature of the optimization, which will make each $G_t$ focus on slightly different parts of the data distribution.
We compare the naive approach to a sophisticated boosting scheme, AdaGAN~\citep{Tolstikhin2017}, where the training samples are re-weighted for each $G_t$ based on discriminator score. The AdaGAN re-weighting puts more focus on the images that the discriminator can easily single out as real, meaning that the current ensemble at a certain ensemble iteration is unable to cover the distribution around these images.
For both ensemble approaches in consideration we use $p_t = 1/T$ for equal contribution from each $G_t$. Moreover, we always keep the number of generated training images fixed to the same number as in the real dataset. That is, if we use $N$ real images, the synthetic dataset from a GAN ensemble with T generators will use $N/T$ images from each $G_t$.

\begin{figure}
\begin{floatrow}
\ffigbox{%
  \includegraphics[width=0.48\textwidth, trim={40pt 10pt 40pt 10pt}]{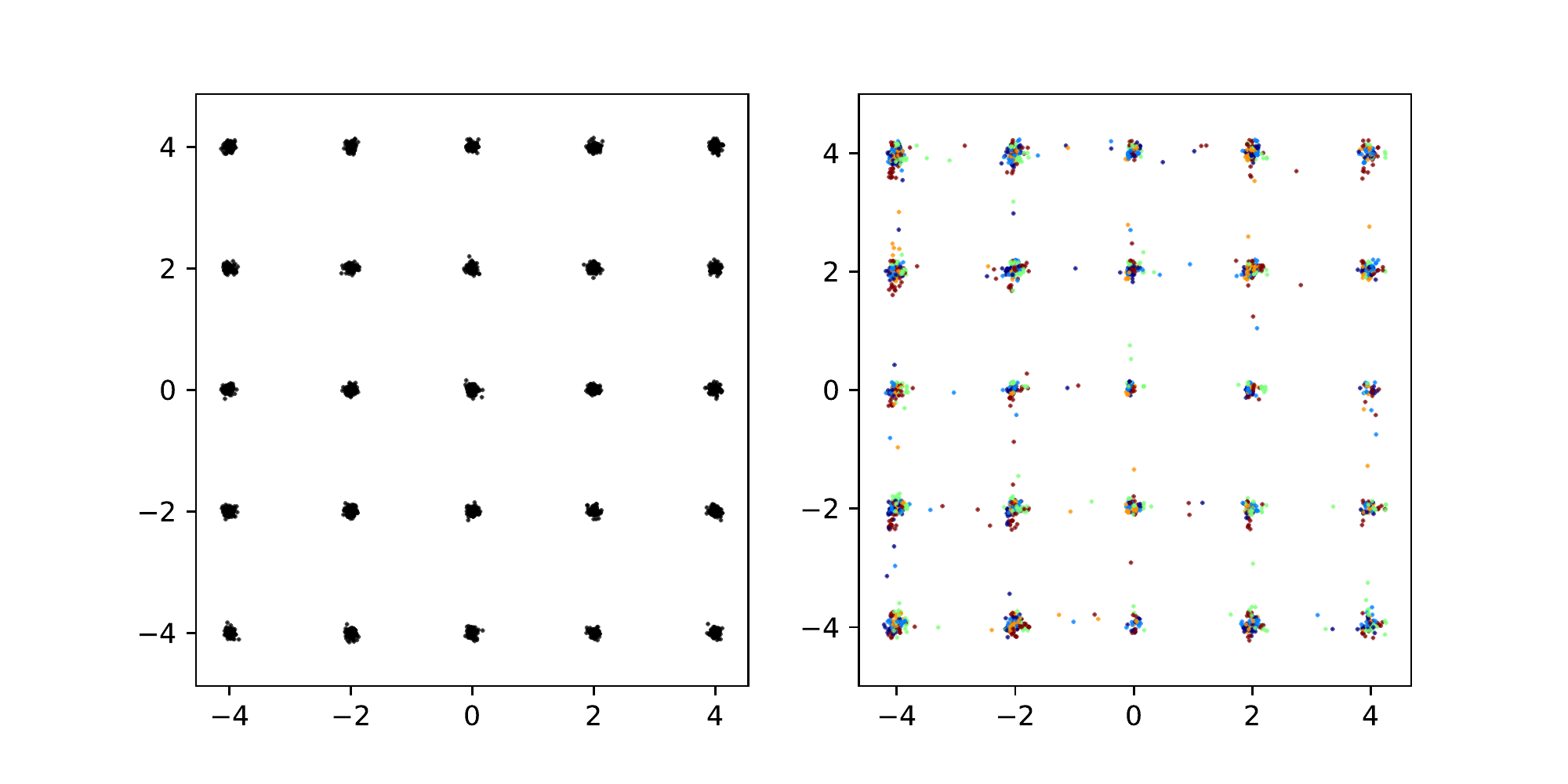}
}{%
  \caption{Ground truth (left) and generated by EnsGAN-5 (right). Colors correspond to different GANs in the ensemble.}%
  \label{fig:2ddata}
}
\capbtabbox{%
    \begin{tabular}{lp{0.13\textwidth}p{0.13\textwidth}}
    & Modes $\;$ (max 25) & High quality samples (\%)  \\
    \hline
    GAN & $18.8\pm2.6$ & $75.7\pm4.6$\\
    PacGAN4 & $24.8\pm0.2$ & $93.6\pm0.6$\\
    MGAN & $22.5\pm0.6$ & $51.1\pm1.6$\\
    ScGAN & $25.0\pm0.0$ & $99.5\pm0.1$\\
    \hline
    EnsGAN-2 & $24.6\pm0.8$ & $75.1\pm3.4$\\
    EnsGAN-3 & $24.8\pm0.7$ & $75.3\pm2.9$\\
    EnsGAN-4 & $25.0\pm0.1$ & $75.2\pm2.6$\\
    EnsGAN-5 & $25.0\pm0.0$ & $75.3\pm2.3$\\
    \end{tabular}
}{%
  \caption{Covered modes and fraction of high quality points, comparing different ensemble sizes to a selection of previous methods.}%
  \label{tab:2ddata}
}
\end{floatrow}
\end{figure}

\heading{Evaluation}
In order to get a sense for the impact of GANs in ensembles on the quality of synthetic training data, we perform a series of experiments on data of increasing difficulty.

%\subsection*{Experiment 1: Synthetic 2D data}
\paragraph{Experiment 1 -- Synthetic 2D data}
As a simple way of analyzing the mode coverage capabilities, we perform an initial experiment on 2D data. We follow the same exact setup as reported in~\citep{Lin2018}, which has been used in other evaluations as well~\citep{Hoang2018,Liu2020}. The dataset is generated by drawing from a mixture of 25 Gaussians arranged in a grid, each representing a separate mode. A generated point is said to be of high quality if it falls within 3 standard deviations from the center of a mode, and a mode is recovered if it has at least one high quality point. We compare to vanilla GAN~\citep{Goodfellow2014} and previously reported numbers for PacGAN~\citep{Lin2018}, MGAN~\citep{Hoang2018}, and self-conditioned GAN (ScGAN)~\citep{Liu2020}. We refer to an ensemble as \emph{EnsGAN-T}, where $T$ is the number of GANs.

The ground truth and generated data are displayed in Figure~\ref{fig:2ddata}. Mode coverage and the fraction of high quality points are presented in Table~\ref{tab:2ddata}. For this dataset 2 or 3 GANs is sufficient to effectively cover all 25 modes. However, while the modes are easily recovered the quality is not affected. We believe that this is not a negative thing since a ``tightening'' around each mode, as can be seen e.g. for ScGAN~\citep{Liu2020}, could potentially affect the diversity although all modes are recovered.

\begin{figure}[t!]
	\centering
	\begin{subfigure}{0.245\linewidth}
		\includegraphics[width=\linewidth, trim={0pt 0pt 0pt 0pt}, clip]{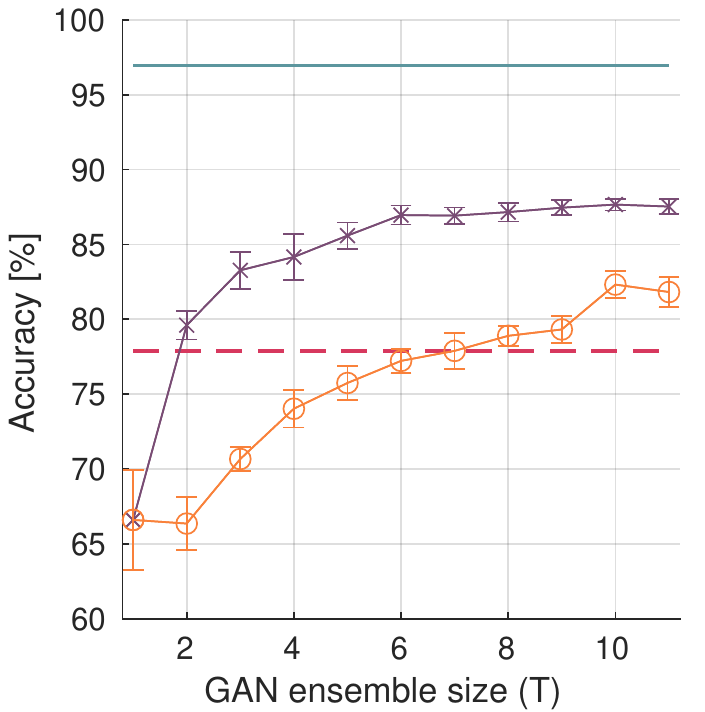}
		\caption{SVHN-II -- DCGAN}
	\end{subfigure}
	\begin{subfigure}{0.245\linewidth}
		\includegraphics[width=\linewidth, trim={0pt 0pt 0pt 0pt}, clip]{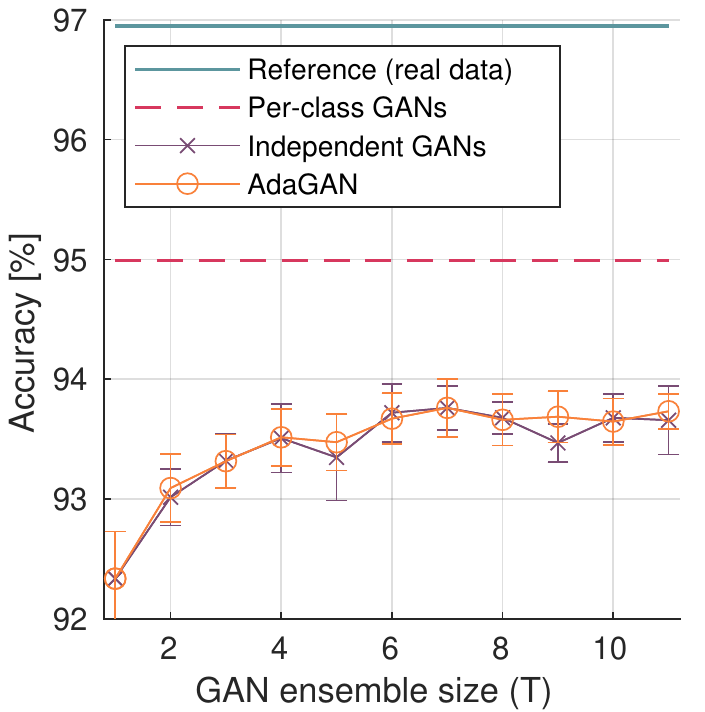}
		\caption{SVHN-II -- PG-GAN}
	\end{subfigure}
	\begin{subfigure}{0.245\linewidth}
		\includegraphics[width=\linewidth, trim={0pt 0pt 0pt 0pt}, clip]{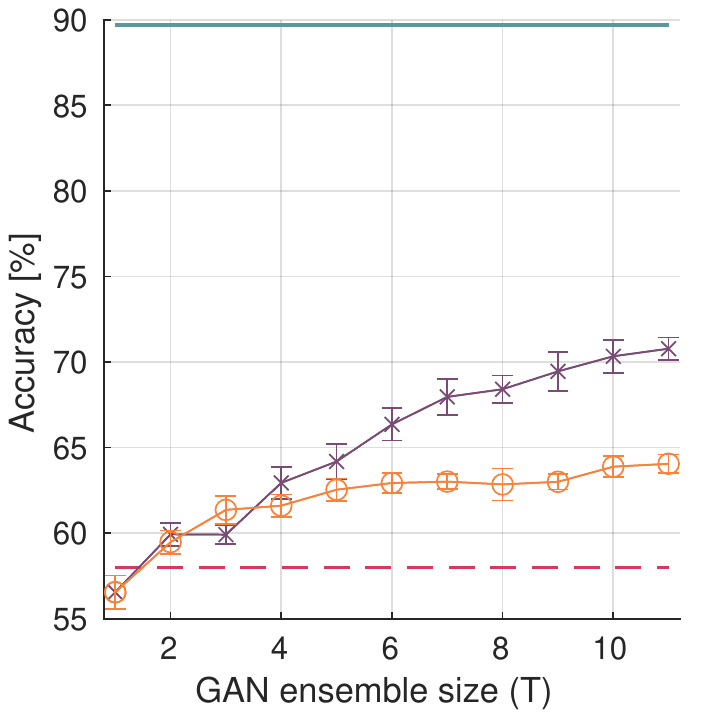}
		\caption{CIFAR-II -- DCGAN}
	\end{subfigure}
	\begin{subfigure}{0.245\linewidth}
		\includegraphics[width=\linewidth, trim={0pt 0pt 0pt 0pt}, clip]{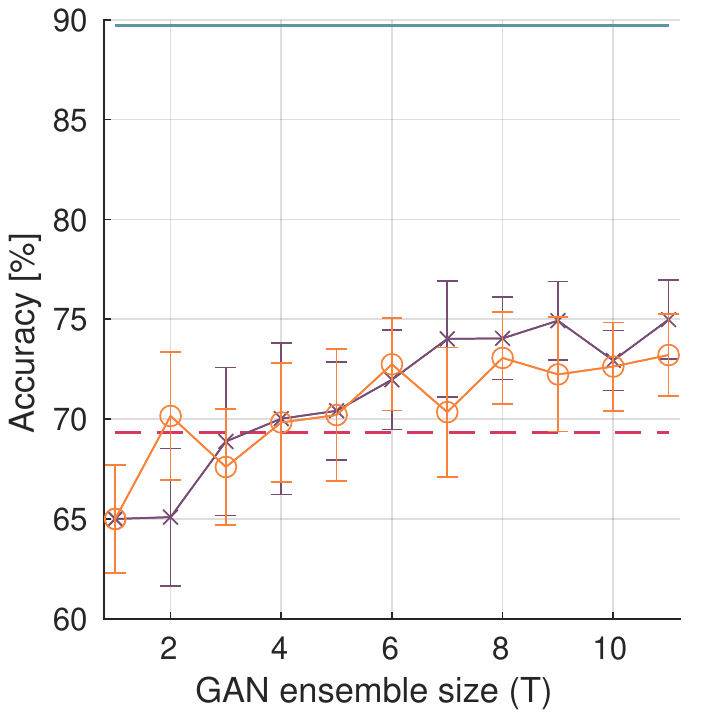}
		\caption{CIFAR-II -- PG-GAN}
	\end{subfigure}
	\caption{\label{fig:svhn_cifar} Classification performance on synthetic datasets, comparing different ensemble sizes and approaches. Each datapoint has been estimated from the mean of 10 separate trainings, and standard deviations are reported with error bars.}
\end{figure}

\begin{figure}[t!]
	\centering
	\begin{subfigure}{0.245\linewidth}
		\includegraphics[width=\linewidth, trim={0pt 0pt 0pt 0pt}, clip]{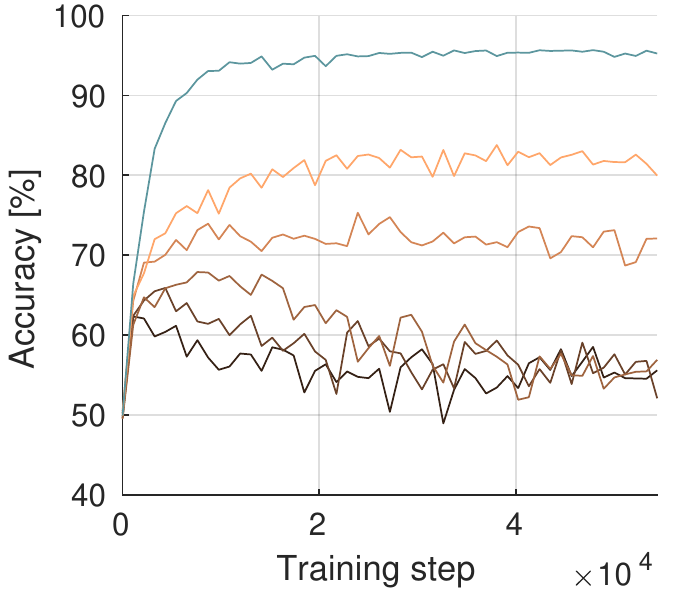}
		\caption{SVHN-II -- DCGAN}
	\end{subfigure}
	\begin{subfigure}{0.245\linewidth}
		\includegraphics[width=\linewidth, trim={0pt 0pt 0pt 0pt}, clip]{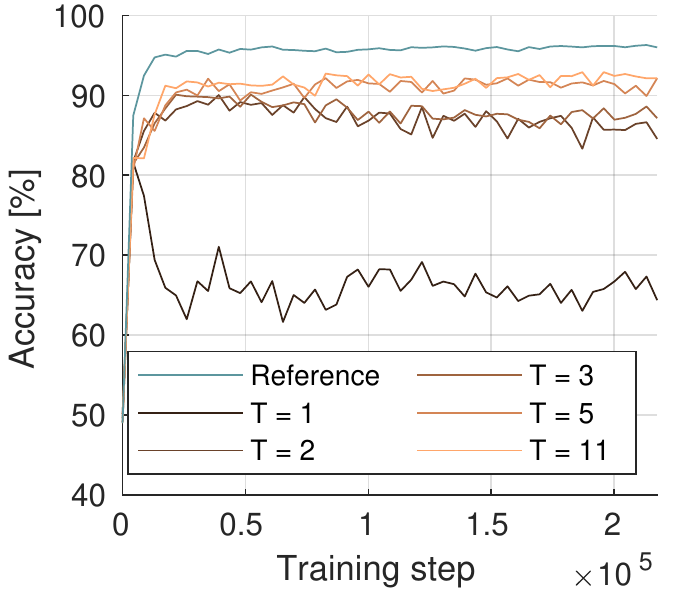}
		\caption{SVHN-II -- PG-GAN}
		\label{fig:trainings_b}
	\end{subfigure}
	\begin{subfigure}{0.245\linewidth}
		\includegraphics[width=\linewidth, trim={0pt 0pt 0pt 0pt}, clip]{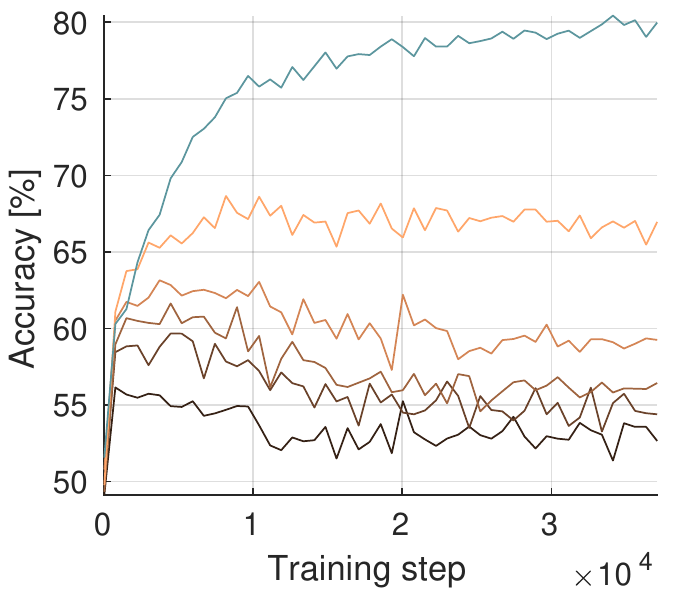}
		\caption{CIFAR-II -- DCGAN}
	\end{subfigure}
	\begin{subfigure}{0.245\linewidth}
		\includegraphics[width=\linewidth, trim={0pt 0pt 0pt 0pt}, clip]{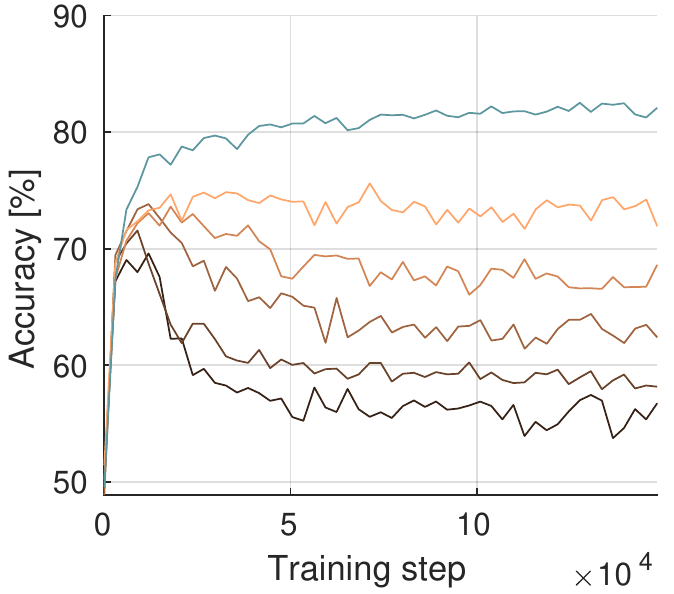}
		\caption{CIFAR-II -- PG-GAN}
	\end{subfigure}
	\caption{\label{fig:trainings} Test accuracy over training steps, for different ensemble sizes. There is a high tendency to overfit with few GANs in an ensemble (small $T$), and the variance between training steps is high.}
	%With more GANs, the generated data becomes more resilient to overfitting and exhibits less variance during training.}
\end{figure}

\paragraph{Experiment 2 -- SVHN/CIFAR-10}
For SVHN~\citep{Netzer2011} and CIFAR-10~\citep{Krizhevsky2009}, each consisting of 10 classes, we partition the data into two separate classes. In SVHN, one class contains digits 0-4 and one uses 5-9. In CIFAR-10 one class is \{\emph{airplane}, \emph{automobile}, \emph{bird}, \emph{cat}, \emph{deer}\}, and the other uses \{\emph{dog}, \emph{frog}, \emph{horse}, \emph{ship}, \emph{truck}\}. We call these two-class datasets SVHN-II and CIFAR-II, respectively. For reference, we also train GANs separately on each of the original 10 classes, followed by partitioning the generated data into the aforementioned classes.
We train ensembles using both DCGAN~\citep{Radford2015} and progressively growing GAN (PG-GAN)~\citep{Karras2018}, which allows us to analyze ensemble performance under different GAN complexities. For details on experimental setup, we refer to the supplementary material.

Figure~\ref{fig:svhn_cifar} shows the accuracy on the downstream classification task, evaluated using a ResNet-18~\citep{He2016}.
%The figure also shows the reference performance from training on the real data, as well as training on a dataset generated by training on the 10 original classes of the dataset separately.
Overall there is a significant improvement when utilizing increasingly large ensembles of GANs, effectively reducing the gap to the performance on real data, especially for DCGAN.
In none of the experiments the more advanced AdaGAN boosting performed significantly better, and for the considered problem independently trained GANs is the better option at least for DCGAN. Moreover, and perhaps the most interesting finding, is how ensembles of DCGAN can outperform a single PG-GAN on CIFAR-II. Since the quality of individual images is not affected by ensembles, this points to how important the improved diversity is for the quality of the synthetic dataset.

To further emphasize the importance of using GANs in ensembles when generating synthetic training data, Figure~\ref{fig:trainings} shows the test performance on real data over training iterations with synthetic data. For both DCGAN and PG-GAN it is problematic to train on synthetic data from a single GAN, experiencing severe problems with over-fitting and stability. Including more models in the ensemble effectively stabilizes the training to better reflect how training on real data behaves.

\paragraph{Experiment 3 -- digital pathology}
As a more challenging scenario, we look at anonymization of the CAMELYON17 dataset~\citep{Litjens2018}. This consists of 50 hematoxylin and eosin (H\&E) stained lymph node whole-slide images containing tumor areas, and was sampled to produce 50K tumor and 50K non-tumor patches at a resolution of 128$\times$128. A separate test set with 15K/15K tumor/non-tumor patches was extracted from slides different from those used to produce the training data. For this dataset, we trained ensembles using PG-GAN, and refer to the supplementary material for details on the experimental setup.

\begin{figure}[t!]
	\centering
	\begin{subfigure}{0.26\linewidth}
		\includegraphics[width=\linewidth, trim={0pt 0pt 0pt 0pt}, clip]{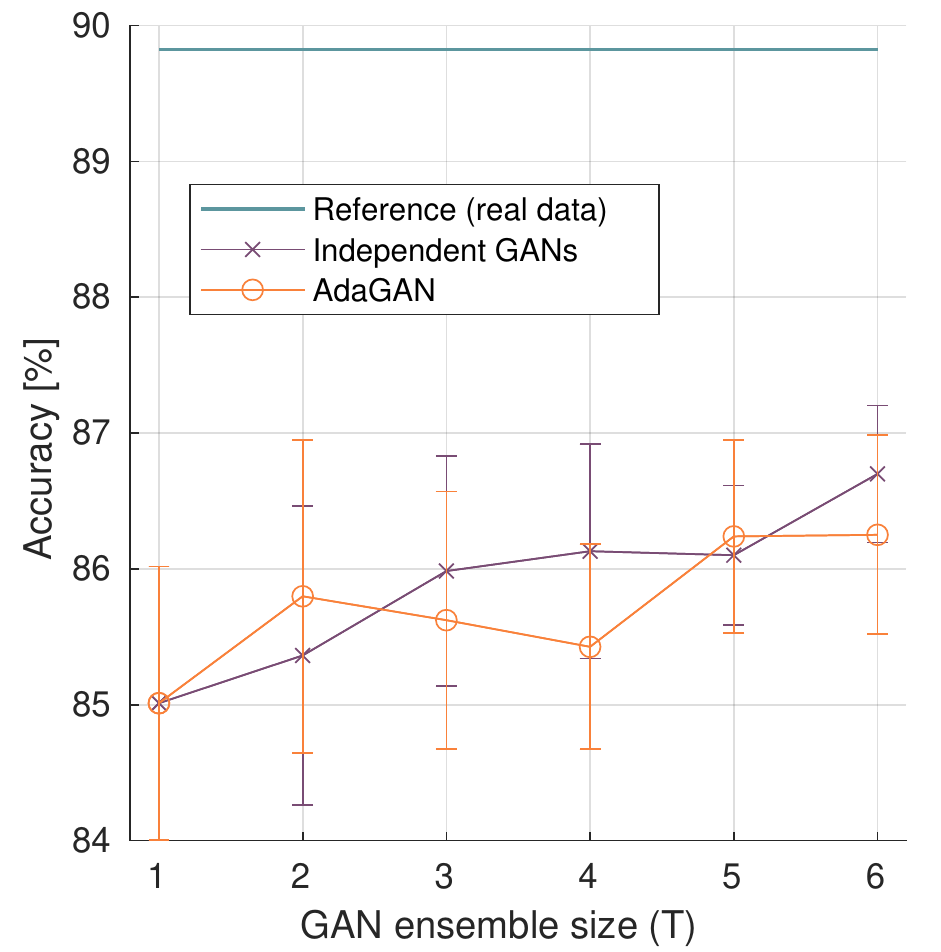}
		\caption{Accuracy, ResNet-18}
	\end{subfigure}
	\hspace{5pt}
	\begin{subfigure}{0.26\linewidth}
		\includegraphics[width=\linewidth, trim={0pt 0pt 0pt 0pt}, clip]{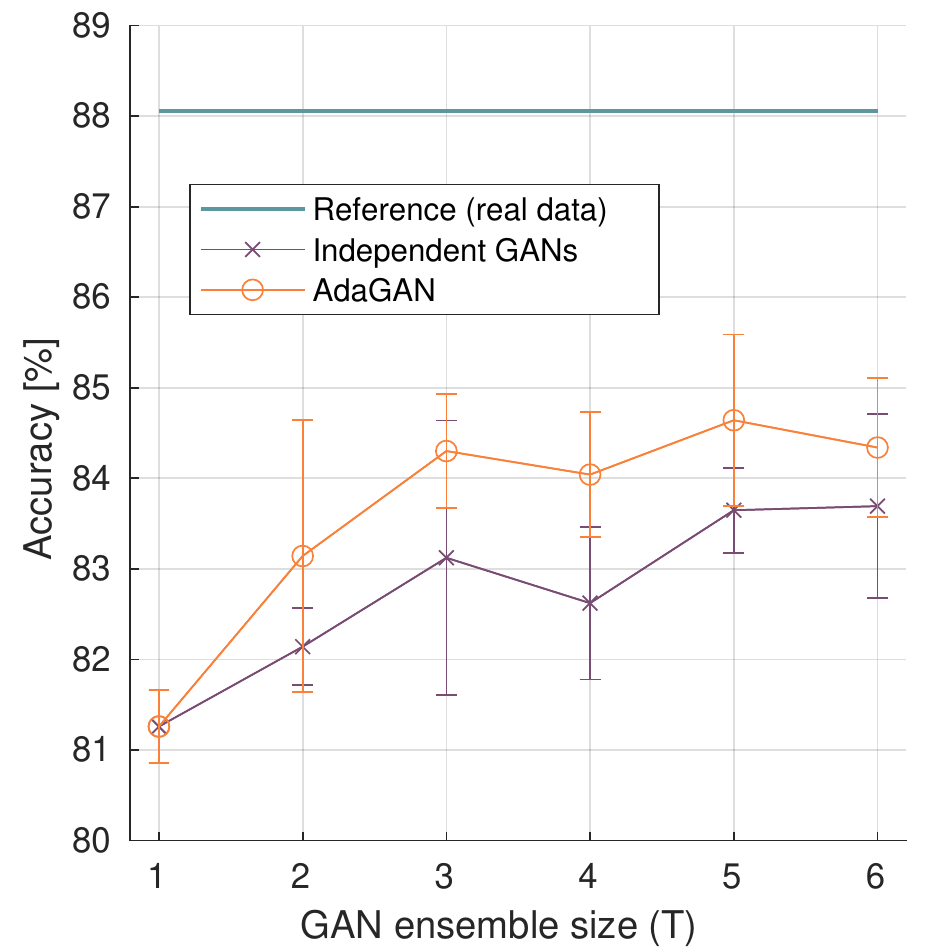}
		\caption{Accuracy, vanilla CNN}
	\end{subfigure}
	\hspace{5pt}
	\begin{subfigure}{0.26\linewidth}
		\includegraphics[width=\linewidth, trim={0pt 0pt 0pt 0pt}, clip]{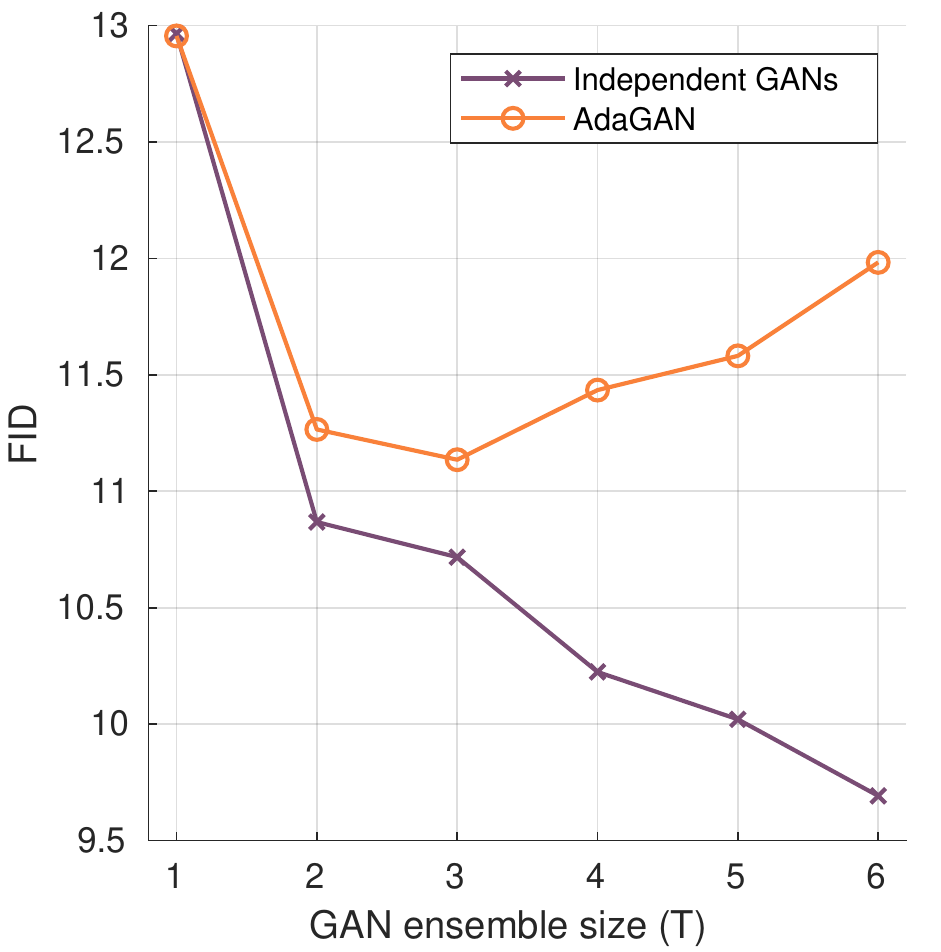}
		\caption{FID}
	\end{subfigure}
	\caption{\label{fig:camelyon} Tumor classification performance on synthetic pathology datasets (a-b), averaged over 10 separate trainings, and FID score evaluated using a model pre-trained on Imagenet (c).}
\end{figure}

Figure~\ref{fig:camelyon} shows the downstream classification performance for a ResNet-18, a vanilla CNN, as well as the Fr\'echet Inception Distance (FID)~\citep{Heusel2017}, for different sizes of ensembles. There are consistent improvements in performance on both classifiers, although most significant for the vanilla CNN, and independent ensembles show a similar performance as AdaGAN.
While there is a large discrepancy in FID between independent ensembles and AdaGAN, this is not reflected in the classification performance, pointing to the weakness of FID as a measure of training dataset quality.

\heading{Discussion and conclusion}
We have provided empirical evidence for the importance of using GANs in ensembles to enable synthetic training data that can be used in isolation from the real data, e.g. in order to act as an anonymization method. The experiments point to how independently trained GANs can be a simple yet powerful technique to improve quality, especially for complex data and simpler GAN models. For a simpler dataset and state-of-the-art GAN, such as PG-GAN on SVHN-II, the impact is less pronounced. However, there is a substantial difference in training behavior when comparing different ensemble sizes (Figure~\ref{fig:trainings_b}). There is also a significant gap to the performance on real data, calling for further work to investigate techniques that can improve GAN synthesized training data. 

One benefit of using independently trained GANs is that this technique does not risk focusing on a too narrow part of the data distribution. This could potentially be a problem with boosted GANs, leading to over-fitting to the exact images used to train the GAN (we refer to the supplementary material for an example), which is not acceptable for anonymization applications.
We have also pointed to how FID is not very representative for training data quality (see Figure~\ref{fig:camelyon} and supplementary material). There has been some studies on training on synthetic data and testing on real data and vice versa~\citep{Shmelkov2018}, which is a better indicator of the dataset quality. Nevertheless, there is a need for formulating standardized quality measures tailored to this specific application of generative models, since the aim most likely differs from many of the existing GAN quality metrics~\citep{Borji2019}. Finally, we see a need for a more comprehensive evaluation of existing techniques for diversifying GANs in the context of synthetic training data generation.

%-- Including training strategies for improving GANs on limited data~\citep{Karras2020}

%For future work, we can see many directions for improving GAN synthesized training data. Self-ensembles could be used in order to improve computational performance~\citep{wang2016ensembles}, and there could be further investigation on boosting strategies that are tailored towards the application of using the generated images for training.
%For real-world scenarios it would also be necessary to further investigate the privacy of GAN generated training data. This could be done by making sure that a generated dataset does not reveal sensitive content, or by placing guarantees on the GAN itself, e.g. by means of differentially private GANs~\citep{Xie2018}.
%Finally, we see a need for a more comprehensive evaluation of existing techniques for diversifying GANs in the context of synthetic training data generation.

\subsubsection*{Acknowledgments}
\vspace{-0.2cm}
%This work was supported by the Wallenberg AI, Autonomous Systems and Software Program (WASP-AI), the research environment ELLIIT, and VINNOVA, grant 2017-02447 (AIDA)

This project was supported by VINNOVA grant 2019-05144 and grant 2017-02447 (AIDA), the Wallenberg Autonomous Systems and Software Program (WASP-NTU and WASP-AI), and the strategic research environment ELLIIT.

\bibliography{iclr2021_conference}
\bibliographystyle{iclr2021_conference}

\newpage
\section*{\Large Supplementary material}
\appendix
\section{2D experiment}
We follow the experimental details provided in~\citep{Lin2018}, which is based on similar experiments in previous evaluations~\citep{Srivastava2017}.
We use the \emph{2D-grid} dataset formulation, which is a mixture of 25 two-dimensional uniform Gaussians with means at the coordinates $(4+2i,4+2j)$ for $i,j \in \{0,...,4\}$ and standard deviation $\sigma = 0.05$. Training is performed by drawing 100K points from this mixture, while metrics are evaluated on 2.5K generated points.

The generator, $G$, uses 4 hidden layers, each with 400 units, and takes as input 2D points drawn from a Gaussian distribution with zero mean and unit variance. Each layer uses batch normalization and ReLU activation. The discriminator, $D$, is composed of 3 hidden layers, each with 200 units, no batch normalization, and uses linear maxout activation with a pool size of 5.
The generator is trained with the loss 
$\mathcal{L}_g = \log\left(1 + e^{D(x_{real})}\right) + \log\left(1 + e^{-D(x_{fake})}\right)$, 
while the discriminator uses 
$\mathcal{L}_d = \log\left(1 + e^{-D(x_{real})}\right) + \log\left(1 + e^{D(x_{fake})}\right)$.
Optimization is performed over 400 epochs, with a minibatch size of 100, using the ADAM optimizer~\citep{Kingma2014}.

\begin{figure}[t]
\includegraphics[width=1.0\linewidth, trim={140pt 10pt 140pt 10pt}]{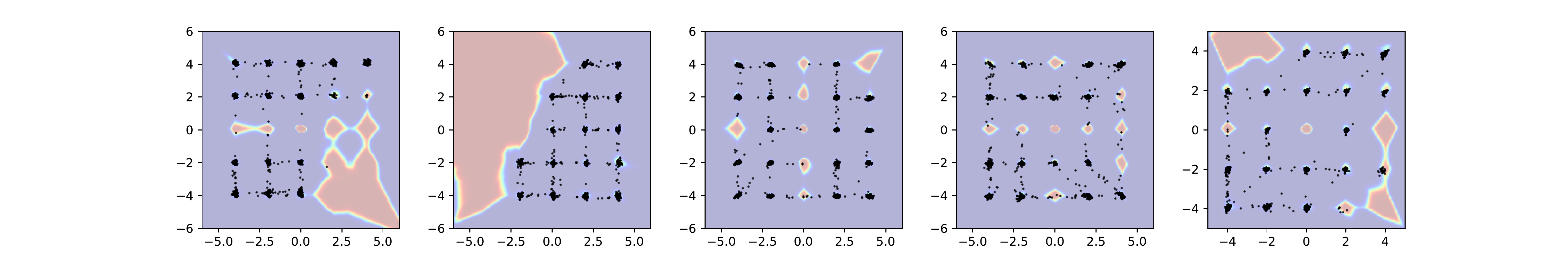}
\caption{5 independently trained GANs on 2D mixtures of Gaussians (2D-grid dataset). The colormap shows discriminator score over the 2D space, highlighting modes that are missed.}
\label{fig:2dgans}
\end{figure}

Using the described data and GAN, the limited mode coverage typically looks like illustrated in Figure~\ref{fig:2dgans}. When combining multiple models, the mode coverage improves quickly, and for this dataset 2-3 models is enough to ensure full mode coverage.

As our focus is on the improvement in mode coverage, the most interesting metric is the number of recovered modes (max 25). A mode is considered recovered if a generated point lands within 3 standard deviations of the mean of the mode. The metric is sensitive to the number of generated points under consideration, but it allows for a direct comparison to previous work. In addition, we also report the fraction of high quality points.

In order to estimate the mean and variance of the recovered modes and fraction of high quality points, we train 25 separate GANs on the 2D dataset and generate 125K points from each model. For each ensemble size $T$, we randomly select $T$ GANs and combine $2500/T$ randomly selected points from each of these, to arrive at a generated dataset with 2500 points. We perform this bootstrapping in 1000 iterations to get a stable estimate of mean and standard deviation.

\section{Image generation experiments}

\paragraph{Datasets}
The datasets with reduced number of classes are formulated with three aspects in mind: 1) increasing the diversity in each class, 2) providing explicit annotations of some main modes of each class, and 3) shortening of the turnaround times since we only have to train 2 GANs for each ensemble iteration instead of 10.

In addition to SVHN-II and CIFAR-II, we also include results for MNIST-II in this supplementary material. The class assignment of each dataset is performed according to:

\begin{table}[h!]
    \centering
    \begin{tabular}{l|l|l}
         Dataset & Class 1 & Class 2 \\
         \hline
         SVHN-II & \emph{0}, \emph{1}, \emph{2}, \emph{3}, \emph{4} & \emph{5}, \emph{6}, \emph{7}, \emph{8}, \emph{9} \\
         CIFAR-II & \emph{airplane}, \emph{automobile}, \emph{bird}, \emph{cat}, \emph{deer} & \emph{dog}, \emph{frog}, \emph{horse}, \emph{ship}, \emph{truck} \\
         MNIST-II & \emph{0}, \emph{1}, \emph{2}, \emph{3}, \emph{4} & \emph{5}, \emph{6}, \emph{7}, \emph{8}, \emph{9} \\
    \end{tabular}
    %\caption{Class assignment of two-class datasets used for experiments}
    \label{tab:data}
\end{table}

\paragraph{GANs}
DCGAN and PG-GAN are used as explained in the original papers~\citep{Radford2015,Karras2018}. DCGAN uses a Tensorflow implementation\footnote{\url{https://github.com/carpedm20/DCGAN-tensorflow}} based on the original PyTorch implementation. PG-GAN uses the original implementation\footnote{\url{https://github.com/tkarras/progressive\_growing\_of\_gans}} provided by the authors.

For each DCGAN, optimization is performed in 50K steps at a batch size of 64, for a total of 3.2M images seen during training. PG-GAN uses different batch-sizes in the growing procedure, but for SVHN-II, CIFAR-II, and MNIST-II there are in total 4M images seen during training, while for CAMELYON17 6.5M images are used for optimization. DCGAN takes around 70-75 min to train on one class of SVHN-II, CIFAR-II or MNIST-II, while PG-GAN takes 6.5-8 hours depending on GPU. For PG-GAN on CAMELYON17 (128$\times$128 pixels resolution), training takes 49-60 hours for one class. We use a combination of Nvidia Titan X, Titan Xp, and GeForce GTX 1080Ti for training.

\paragraph{Boosting on SVHN-II, CIFAR-II and MNIST-II}
For the discriminator used for boosting with AdaGAN, the CNN is composed of three convolutional layers with 16, 32, and 64 channels. Filter size is set to 3$\times$3 for all layers. Each layer is followed by 2$\times$2 max-pooling and batch normalization. The convolutional layers are followed by two fully connected layers, with 64/64 and 64/2 input/output units, respectively. The first fully connected layer is followed by dropout with probability 0.75. Throughout the network, ReLU activation is used.
The network is initialized with the Glorot random scheme~\citep{Glorot2010}, and then optimized for 4 epochs using the ADAM optimizer~\citep{Kingma2014} ($\beta_1=0.9$, $\beta_2=0.999$), starting at learning rate $5\times10^{-3}$ and decayed down to $3.8\times10^{-4}$ during the course of training.

\paragraph{Boosting on CAMELYON17}
When training on CAMELYON17, the boosting network and training is the same as above, except for the following. Two extra convolutional layers are used, both with 64 channels, and followed by max-pooling. The network is trained for 2 epochs, with a learning rate starting at $1\times10^{-3}$ and decayed down to $7.7\times10^{-5}$. As the dataset is larger, 2 epochs corresponds to the same number of training steps as 4 epochs on e.g. CIFAR-II. Moreover, there was a strong tendency to overfit, which was alleviated by reducing the learning rate, and applying augmentations. The augmentations include random vertical and horizontal flips, as well as some random changes in brightness, contrast, hue, and saturation.

\paragraph{Classification evaluation}
Evaluation in terms of classification performance when training on the synthetic datasets is performed using ResNet-18\footnote{\url{https://github.com/tensorpack/tensorpack/tree/master/examples/ResNet}}~\citep{He2016}. In this supplementary material we also show results using a vanilla CNN, which is the same model used for boosting and described above. We term this \emph{SimpleCNN} in the results.
Performance is evaluated on the test set from the real data, i.e. images not used for training of the GANs. With few GANs, i.e. small or no ensemble used, a significant problem is over-fitting of the classifier (see Figure~\ref{fig:trainings}). This makes it problematic to extract a validation set from the training data, for defining an early-stopping criterion. Instead, we opted to use the test data as a validation set, and simply used the best validation performance as evaluation measure. This scenario would correspond to a real-world situation, where you get synthetic data for training and have a smaller set of real validation images to optimize performance over. Also, this scenario is the most difficult in terms of demonstrating improvements with ensembles; since the over-fitting of small ensembles is less penalized with this strategy, there will be a smaller discrepancy between small and large ensembles.

\begin{figure}[t!]
	\centering
	\begin{subfigure}{0.325\linewidth}
		\includegraphics[width=\linewidth, trim={0pt 0pt 0pt 0pt}, clip]{fig/gansembles/svhn-10-2_dcgan_resnet.pdf}
		\caption{SVHN-II -- DCGAN}
	\end{subfigure}
	\begin{subfigure}{0.325\linewidth}
		\includegraphics[width=\linewidth, trim={0pt 0pt 0pt 0pt}, clip]{fig/gansembles/cifar-10-2_dcgan_resnet.pdf}
		\caption{CIFAR-II -- DCGAN}
	\end{subfigure}
	\begin{subfigure}{0.325\linewidth}
		\includegraphics[width=\linewidth, trim={0pt 0pt 0pt 0pt}, clip]{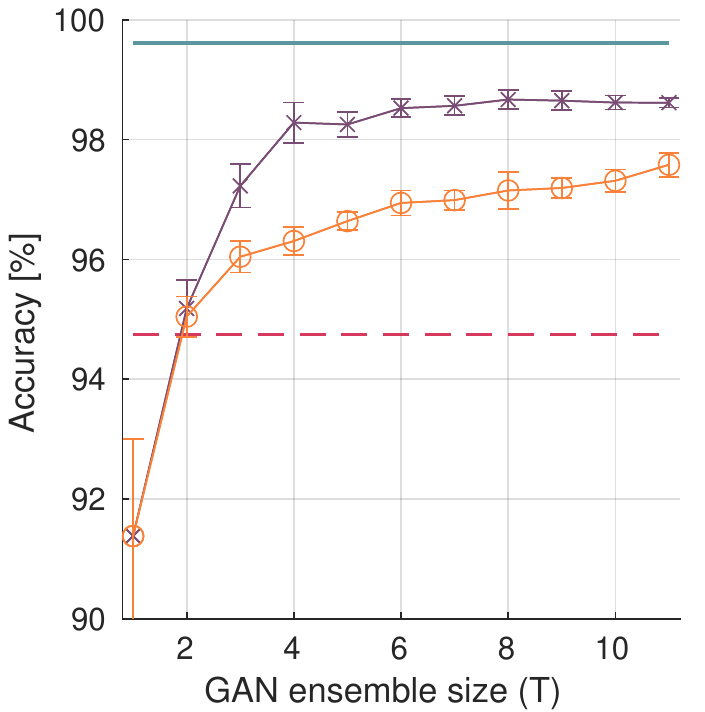}
		\caption{MNIST-II -- DCGAN}
	\end{subfigure}
	
	\begin{subfigure}{0.325\linewidth}
		\includegraphics[width=\linewidth, trim={0pt 0pt 0pt 0pt}, clip]{fig/gansembles/svhn-10-2_pgan_resnet.pdf}
		\caption{SVHN-II -- PG-GAN}
	\end{subfigure}
	\begin{subfigure}{0.325\linewidth}
		\includegraphics[width=\linewidth, trim={0pt 0pt 0pt 0pt}, clip]{fig/gansembles/cifar-10-2_pgan_resnet.pdf}
		\caption{CIFAR-II -- PG-GAN}
	\end{subfigure}
	\begin{subfigure}{0.325\linewidth}
		\includegraphics[width=\linewidth, trim={0pt 0pt 0pt 0pt}, clip]{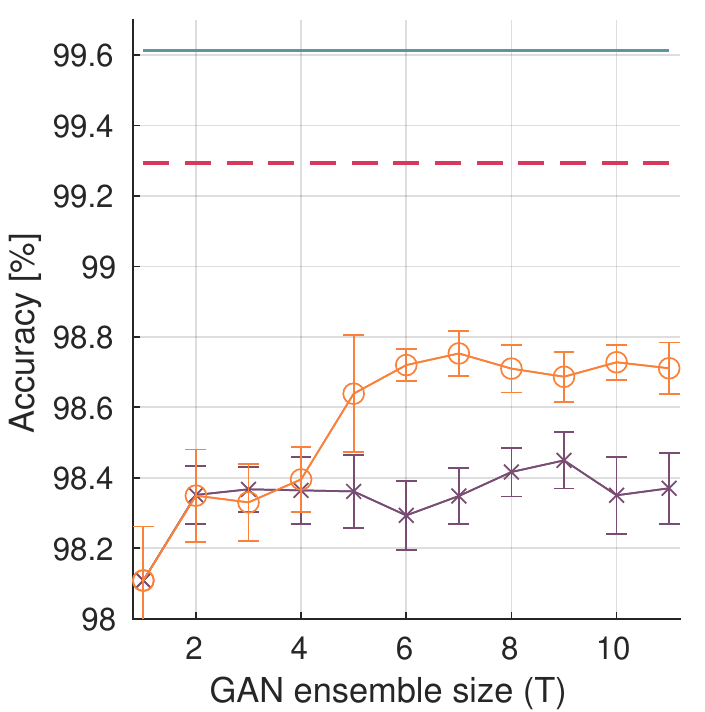}
		\caption{MNIST-II -- PG-GAN}
	\end{subfigure}
	\caption{\label{fig:gansemble_resnet} ResNet-18 classification performance on synthetic datasets generated by different ensemble methods and ensemble sizes. Sub-figures show the performance for different combinations of GAN and dataset. Each data point has been estimated as the mean over 10 separate classification trainings, and the error bars provide standard deviations. The horizontal lines illustrate reference performance on real data, and training GANs separately on each of the 10 classes of the original 10-class datasets (as compared to the 2-class versions of the data used for training ensembles). The legends in (d) apply to all sub-figures.}
\end{figure}

\begin{figure}[t!]
	\centering
	\begin{subfigure}{0.325\linewidth}
		\includegraphics[width=\linewidth, trim={0pt 0pt 0pt 0pt}, clip]{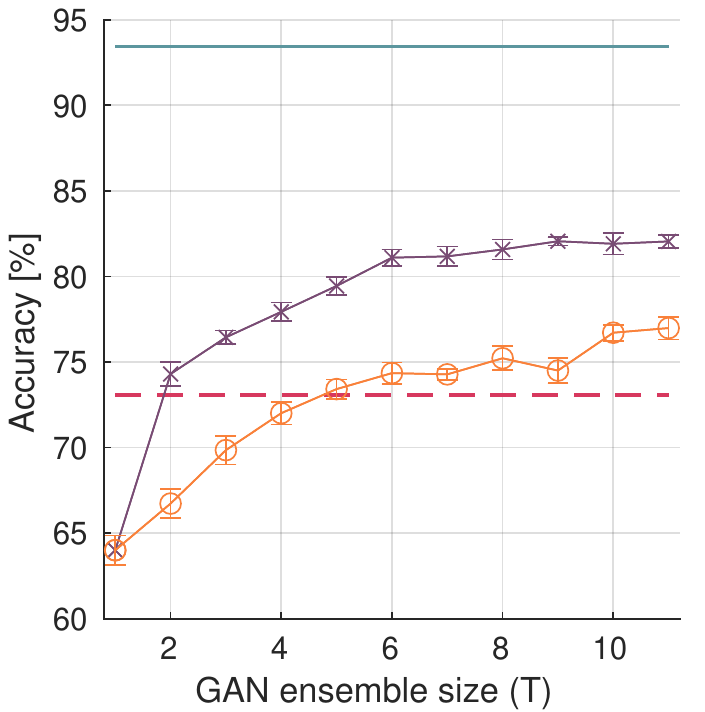}
		\caption{SVHN-II -- DCGAN}
	\end{subfigure}
	\begin{subfigure}{0.325\linewidth}
		\includegraphics[width=\linewidth, trim={0pt 0pt 0pt 0pt}, clip]{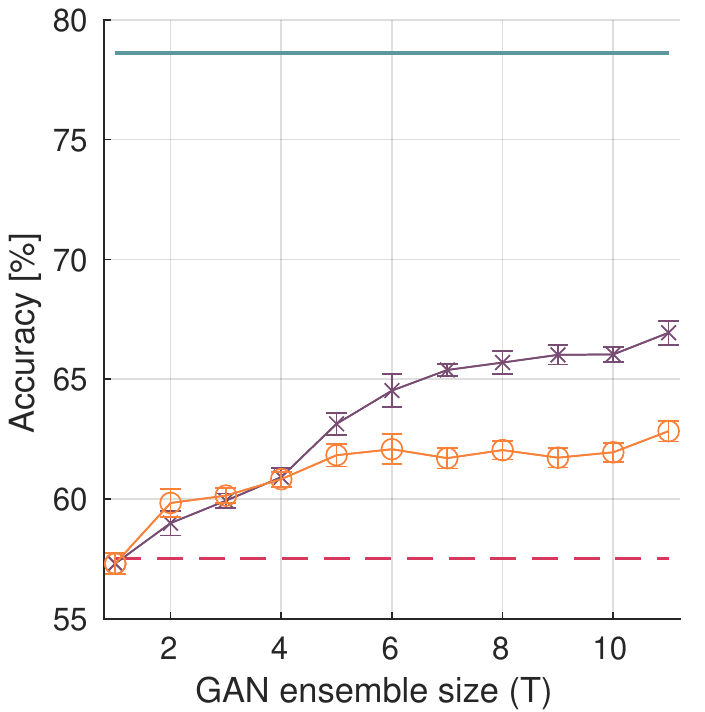}
		\caption{CIFAR-II -- DCGAN}
	\end{subfigure}
	\begin{subfigure}{0.325\linewidth}
		\includegraphics[width=\linewidth, trim={0pt 0pt 0pt 0pt}, clip]{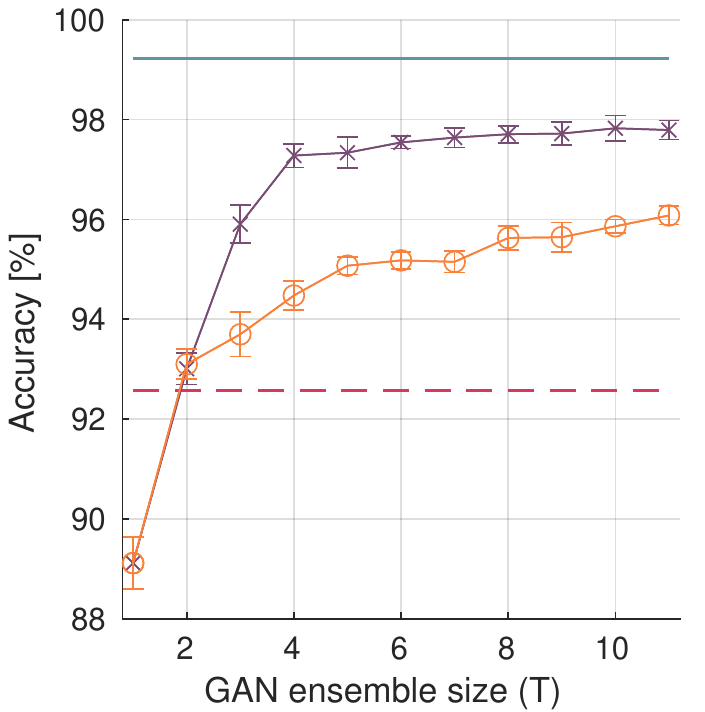}
		\caption{MNIST-II -- DCGAN}
	\end{subfigure}
	
	\begin{subfigure}{0.325\linewidth}
		\includegraphics[width=\linewidth, trim={0pt 0pt 0pt 0pt}, clip]{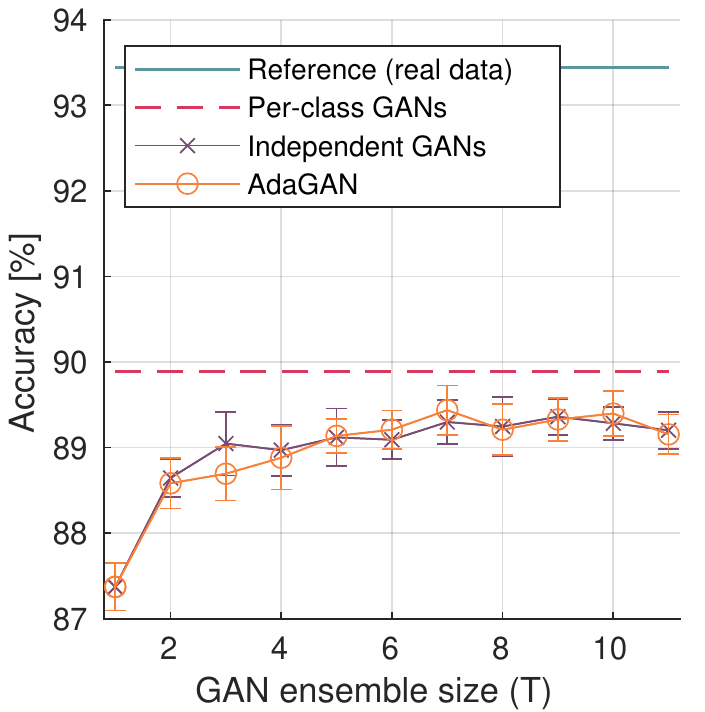}
		\caption{SVHN-II -- PG-GAN}
	\end{subfigure}
	\begin{subfigure}{0.325\linewidth}
		\includegraphics[width=\linewidth, trim={0pt 0pt 0pt 0pt}, clip]{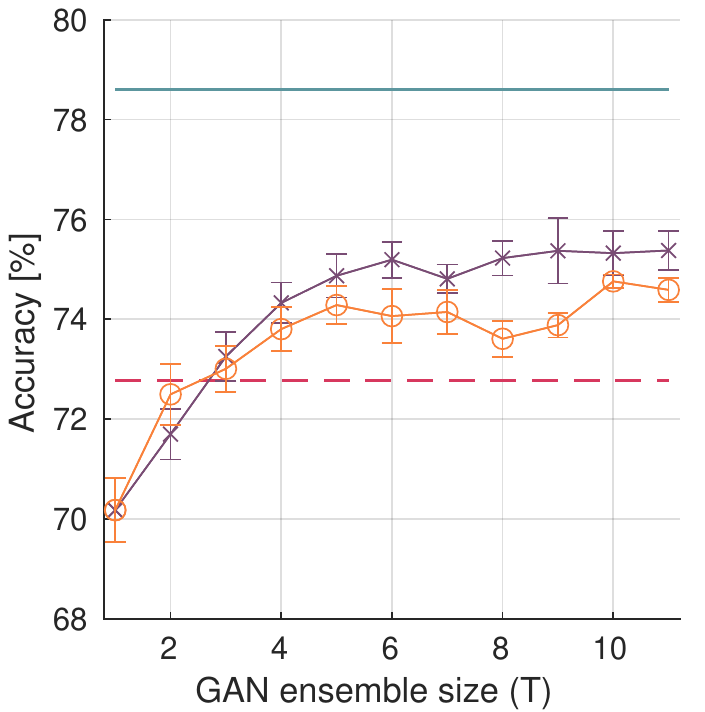}
		\caption{CIFAR-II -- PG-GAN}
	\end{subfigure}
	\begin{subfigure}{0.325\linewidth}
		\includegraphics[width=\linewidth, trim={0pt 0pt 0pt 0pt}, clip]{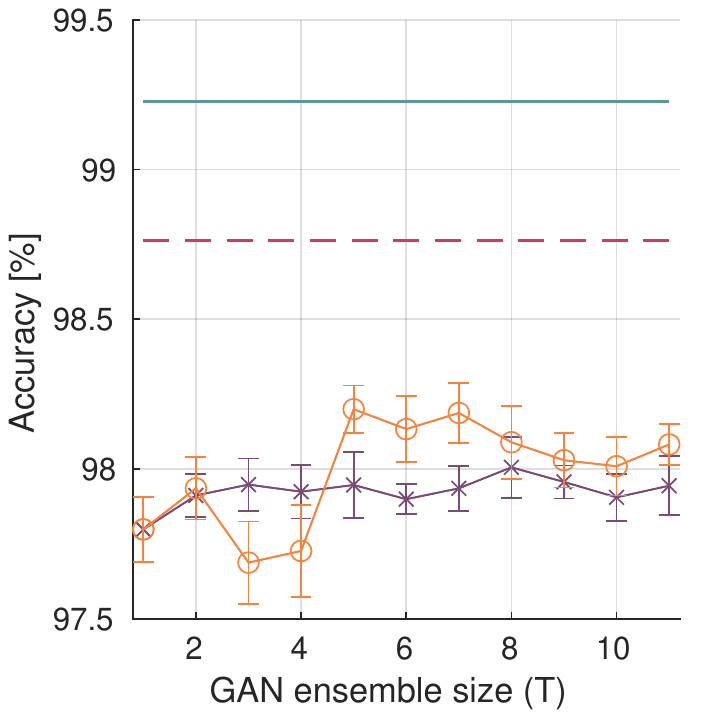}
		\caption{MNIST-II -- PG-GAN}
	\end{subfigure}
	\caption{\label{fig:gansemble_simple} Same as Figure~\ref{fig:gansemble_resnet}, but using SimpleCNN instead of ResNet-18.}
\end{figure}

\begin{figure}[t!]
	\centering
	\begin{subfigure}{0.325\linewidth}
		\includegraphics[width=\linewidth, trim={0pt 0pt 0pt 0pt}, clip]{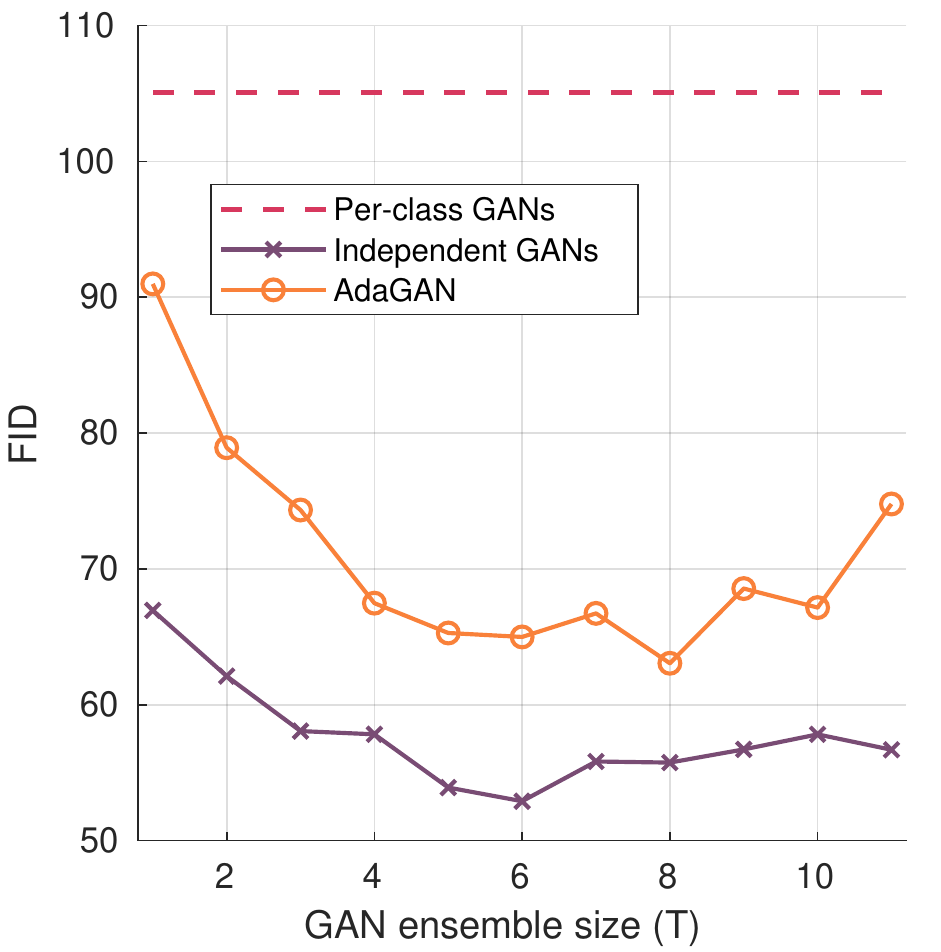}
		\caption{SVHN-II -- DCGAN}
	\end{subfigure}
	\begin{subfigure}{0.325\linewidth}
		\includegraphics[width=\linewidth, trim={0pt 0pt 0pt 0pt}, clip]{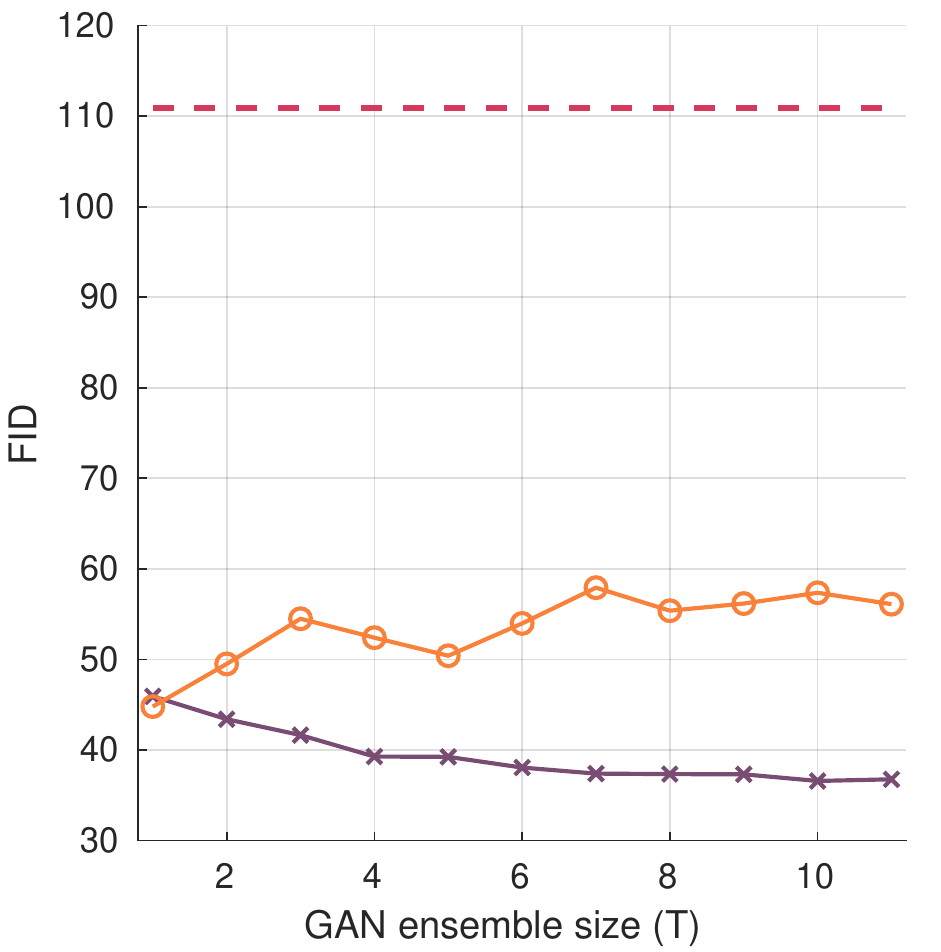}
		\caption{CIFAR-II -- DCGAN}
	\end{subfigure}
	\begin{subfigure}{0.325\linewidth}
		\includegraphics[width=\linewidth, trim={0pt 0pt 0pt 0pt}, clip]{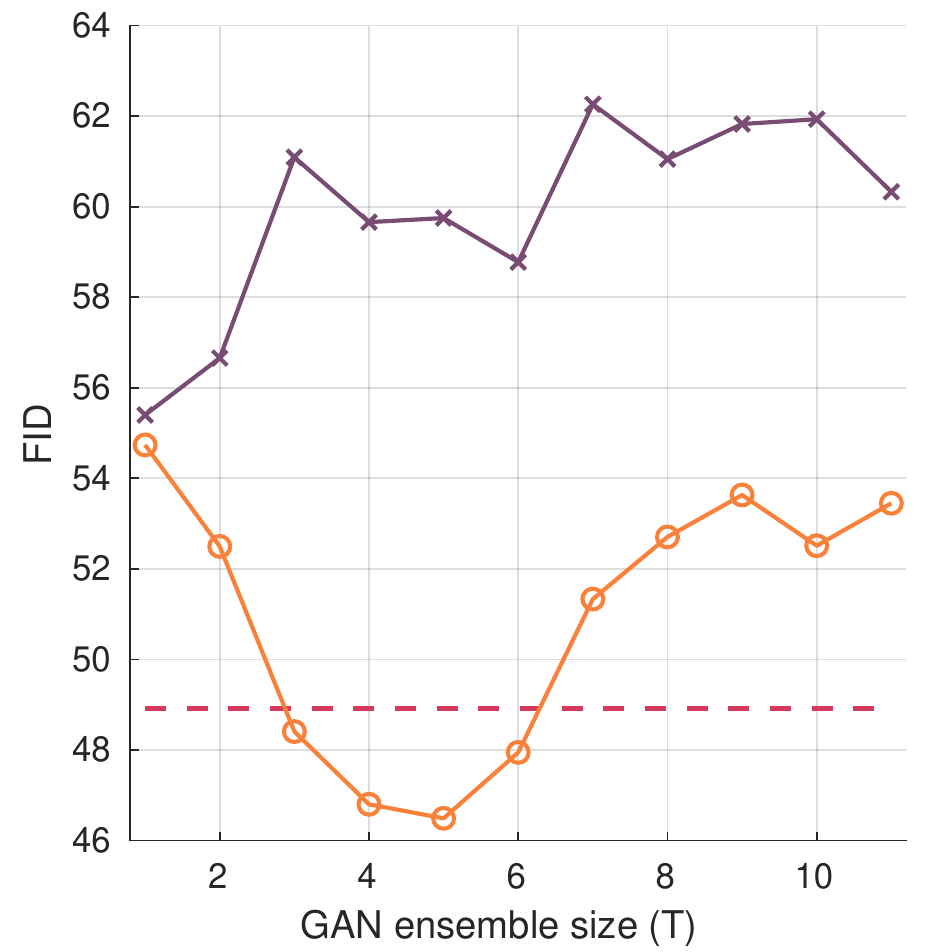}
		\caption{MNIST-II -- DCGAN}
	\end{subfigure}
	
	\begin{subfigure}{0.325\linewidth}
		\includegraphics[width=\linewidth, trim={0pt 0pt 0pt 0pt}, clip]{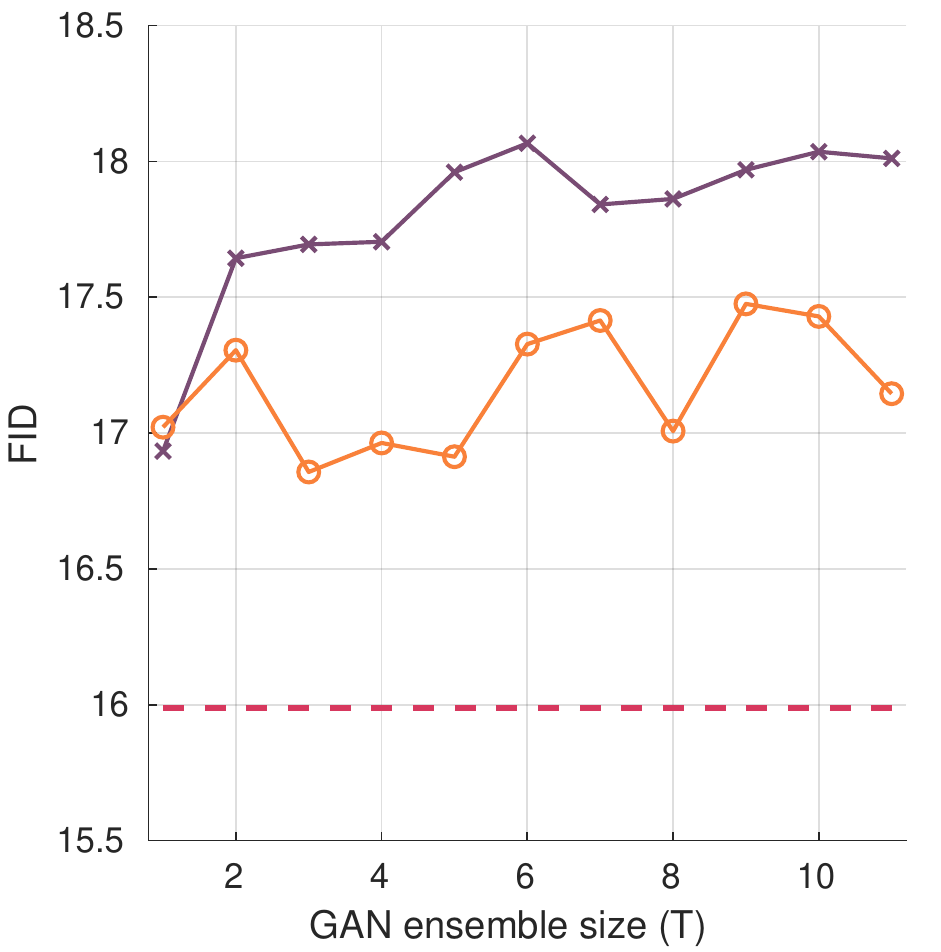}
		\caption{SVHN-II -- PG-GAN}
	\end{subfigure}
	\begin{subfigure}{0.325\linewidth}
		\includegraphics[width=\linewidth, trim={0pt 0pt 0pt 0pt}, clip]{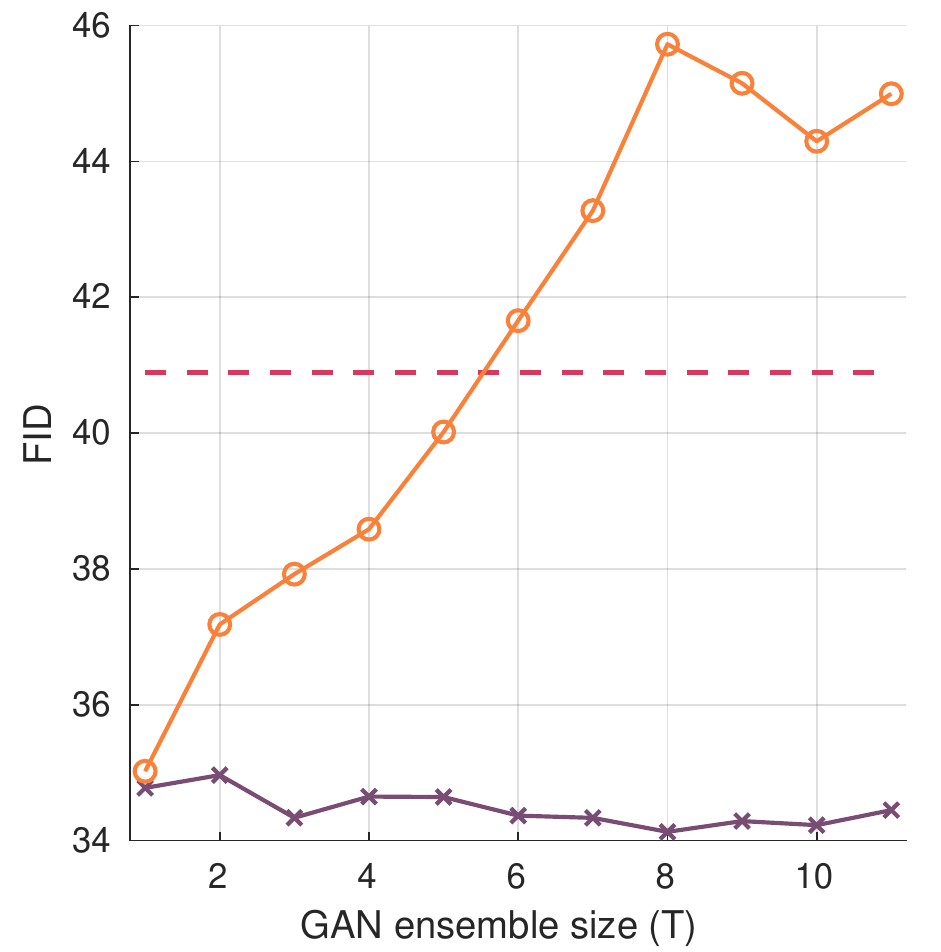}
		\caption{CIFAR-II -- PG-GAN}
	\end{subfigure}
	\begin{subfigure}{0.325\linewidth}
		\includegraphics[width=\linewidth, trim={0pt 0pt 0pt 0pt}, clip]{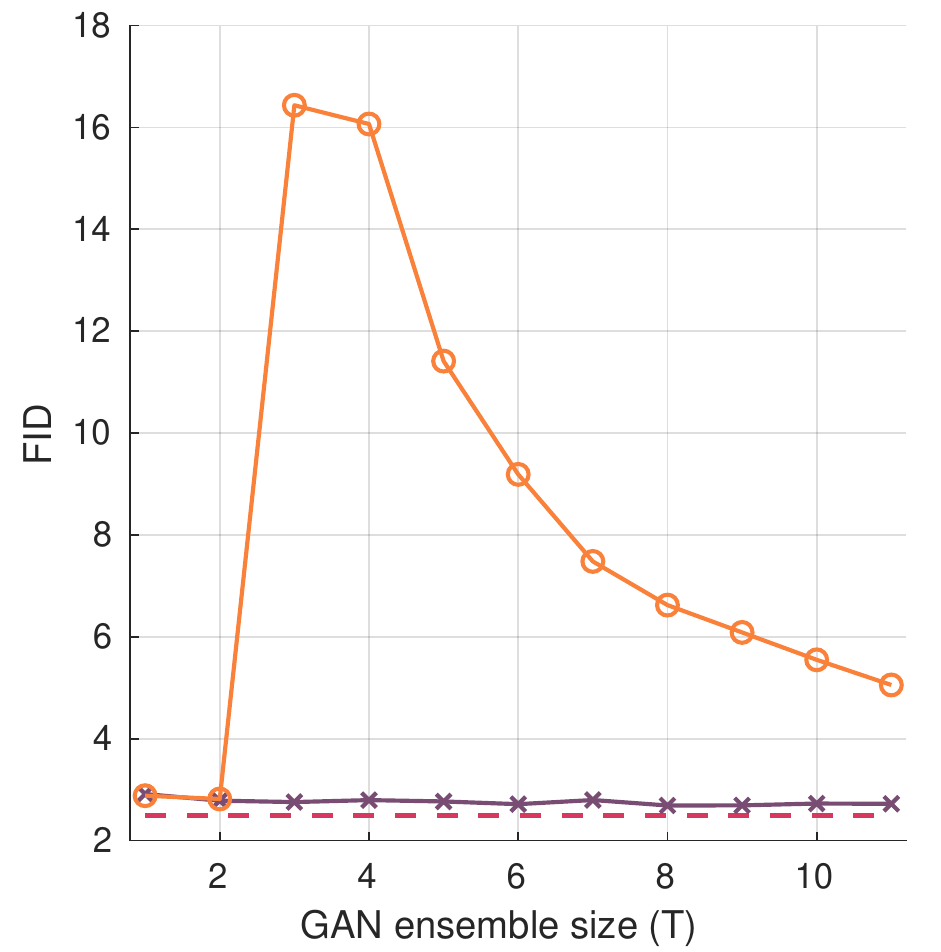}
		\caption{MNIST-II -- PG-GAN}
	\end{subfigure}
	\caption{\label{fig:fid} FID evaluation on synthetic datasets.}
\end{figure}

\section{Supplementary results}
\paragraph{Classification performance}
The full set of evaluations using classification performance on SVHN-II, CIFAR-II and MNIST-II is shown in Figure~\ref{fig:gansemble_resnet} and~\ref{fig:gansemble_simple}, using ResNet-18 and SimpleCNN, respectively. The performance has been averaged over 10 individual training runs, and the standard deviations are shown as error bars. Generally, there is the same trends with both classifier models. Synthetic training data produced by simpler GANs (DCGAN) benefit more from ensembles, as well as less simplistic datasets (CIFAR-II). Using PG-GAN on a simpler dataset (SVHN-II or MNIST-II) also yields noticeable improvements, but of less magnitude compared to the other examples. 

\paragraph{FID}
The Fr\'echet Inception Distance (FID)~\citep{Heusel2017} has been computed for all synthetic datasets, by means of a CNN trained on Imagenet, and Figure~\ref{fig:fid} shows the results. Comparing to Figure~\ref{fig:gansemble_resnet} and~\ref{fig:gansemble_simple} it is clear how FID is not sufficient for evaluating the quality of a synthetic dataset intended for training deep classifiers. Generally, FID could be lowered by ensembled DCGANs, but for PG-GAN there is only a small decrease on CIFAR-II. With boosted GANs there is a stronger tendency to increase FID as compared to independently trained GANs.

\begin{figure}[t!]
	\centering
	\begin{subfigure}{0.49\linewidth}
		\includegraphics[width=\linewidth, trim={0pt 0pt 0pt 0pt}, clip]{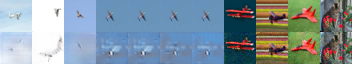}
		\caption{Class 1: \{\emph{airplane}, \emph{automobile}, \emph{bird}, \emph{cat}, \emph{deer}\}}
	\end{subfigure}
	\hspace{3pt}
	\begin{subfigure}{0.49\linewidth}
		\includegraphics[width=\linewidth, trim={0pt 0pt 0pt 0pt}, clip]{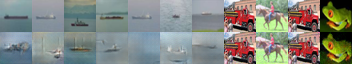}
		\caption{Class 2: \{\emph{dog}, \emph{frog}, \emph{horse}, \emph{ship}, \emph{truck}\}}
	\end{subfigure}
	\caption{\label{fig:nn_adagan} Nearest neighbors between real (top) and synthetic (bottom) datasets, for different ensemble sizes (left to right). The datasets have been generated by means of DCGAN on CIFAR-10, using the AdaGAN boosting scheme. For this setup AdaGAN can experience over-fitting when the number of boosting iterations increase, as seen in the direct replication of training images after 7 iterations of boosting.}
\end{figure}

\begin{figure}[t!]
	\centering
	\begin{subfigure}{0.49\linewidth}
		\includegraphics[width=\linewidth, trim={0pt 0pt 0pt 0pt}, clip]{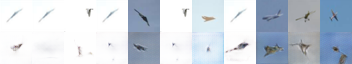}
		\caption{Class 1: \{\emph{airplane}, \emph{automobile}, \emph{bird}, \emph{cat}, \emph{deer}\}}
	\end{subfigure}
	\hspace{3pt}
	\begin{subfigure}{0.49\linewidth}
		\includegraphics[width=\linewidth, trim={0pt 0pt 0pt 0pt}, clip]{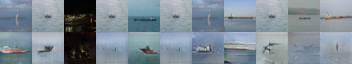}
		\caption{Class 2: \{\emph{dog}, \emph{frog}, \emph{horse}, \emph{ship}, \emph{truck}\}}
	\end{subfigure}
	\caption{\label{fig:nn_rand} Same as Figure~\ref{fig:nn_adagan}, but for independent ensembles (no boosting). There are no obvious examples of direct replication of training images.\vspace{13cm}}
\end{figure}

\paragraph{Over-fitting}
When training GANs on small datasets there is a risk that the generator will over-fit to some of the exact training images. In an application such as anonymization this can not be accepted. The advantage of independently trained GANs in an ensemble is that each GAN is trained on the full dataset, so that the risk of over-fitting is not increased as compared to using a single GAN. For boosted GANs, however, the risk could increase with the number of iterations since the boosting scheme will focus on a narrowed data distribution. To illustrate this problem, Figure~\ref{fig:nn_adagan} and~\ref{fig:nn_rand} show the nearest neighbors between real and synthetic datasets. The neighbors have been found through an exhaustive search of all combinations of real and synthetic images, and measured using the mean squared pixel-wise difference between images.
In Figure~\ref{fig:nn_adagan}, it is evident how training images are starting to be replicated after 7 iterations of boosting. For the independently trained GANs (Figure~\ref{fig:nn_rand}), and in all experiments using PG-GAN, we did not see such problems with over-fitting.

\end{document}